\documentclass[letterpaper,11pt]{extarticle}
 
\usepackage{breakcites}
\usepackage{hyperref}       
\usepackage{url}            
\usepackage{booktabs}       
\usepackage{amsfonts}       
\usepackage{amsmath}
\usepackage{mathtools}
\usepackage{nicefrac}       
\usepackage{microtype}      
\usepackage{lipsum}
\usepackage{graphicx}
\usepackage{wrapfig}
\usepackage[left=1in,right=1in,top=1.5in,bottom=1.8in]{geometry}
\usepackage{xcolor}
\usepackage{setspace} 
\usepackage{subcaption}
\usepackage{caption}
\usepackage{gensymb}
\usepackage[]{algorithm2e}
\usepackage[font=small,labelfont=bf]{caption}
\definecolor{saycolor}{HTML}{21618C}

\usepackage[normalem]{ulem}  
\usepackage{advdate}

\title{Deep Gaussian Processes: A Survey}
 
\begin{document}

\author{
Kalvik Jakkala\\
Department of Computer Science \\
University of North Carolina at Charlotte\\
Charlotte, NC, USA \\
\texttt{kjakkala@uncc.edu}
}

\date{}

\maketitle

\begin{abstract}
Gaussian processes are one of the dominant approaches in Bayesian learning. Although the approach has been applied to numerous problems with great success, it has a few fundamental limitations. Multiple methods in literature have addressed these limitations. However, there has not been a comprehensive survey of the topics as of yet. Most existing surveys focus on only one particular variant of Gaussian processes and their derivatives. This survey details the core motivations for using  Gaussian processes, their mathematical formulations, limitations, and research themes that have flourished over the years to address said limitations.

Furthermore, one particular research area is Deep Gaussian Processes (DGPs), it has improved substantially in the past decade. The significant publications that advanced the forefront of this research area are outlined in their survey. Finally, a brief discussion on open problems and research directions for future work is presented at the end. 
\end{abstract}

\section{Introduction}
There have been numerous advances in the field of machine learning in recent years. Most of these advances can be attributed to backpropagation, large datasets, and computational resource improvements. However, most of the currently popular machine learning approaches mainly, deep learning methods, are based on frequentist approaches, which entail making any prediction decisions by studying the correlations among features and predictions in a dataset. The problem with such an approach is that it is easy to overfit to the available dataset and risk learning unwanted biasing in the dataset. 

Furthermore, current approaches make it difficult and unintuitive to introduce any prior domain knowledge into a prediction model. Several real-world problems have domain experts; incorporating their knowledge could result in substantially better models. However, most deep learning approaches do not accommodate such incorporations and require application-specific methods to be developed to address such an issue.

Prediction uncertainty is an important metric that needs to be estimated by reliable models. Most data sources contain non-negligible quantities of noise that might hinder the performance of a prediction model. It is also not uncommon to have test data samples that do not closely resemble the training dataset distribution. In such cases, it is essential to know the model's prediction uncertainty. If the model were to be used in a mission-critical task without accounting for its prediction uncertainty, it could lead to catastrophic results. 

Another major drawback of conventional deep learning approaches is model comparison. Deep learning approaches are parametric and require explicit definitions for the model architecture. Moreover, model architectures are application specific. Often multiple model architectures need to be compared against each other to determine which is the best model for a task. However, it is nontrivial to factor in model size in terms of its parameter count and accuracy for comparison. 

The limitations mentioned above are addressed by Bayesian approaches with varying degrees of ease and efficiency. We can incorporate domain knowledge with a prior distribution, prediction uncertainty can be estimated with prediction variance, and models can be aptly compared against each other with the Bayes factor.

Apart from the advantages mentioned above, another interesting feature of Bayesian methods is that they facilitate causal modeling of any system or process. Indeed, most classification or regression problems require a chain of sub-decisions, each of which would lead to the final prediction. However, conventional deep learning approaches are not particularly amenable to specifying such causal models. The Bayesian framework, along with do-calculus \cite{Pearl00, PearlM18}, can be used to specify such structures in models. 

The advantages of Bayesian approaches raise the question of why they dont have widespread adaptation yet.  Bayesian approaches often incur heavy computational expense or outright intractabilities making them infeasible for several problems. Nonetheless, these methods are steeped with history and have been used to solve numerous problems with substantial ramifications \cite{Mcgrayne11}. Time and time again, the Bayesian framework has proved itself to be worthy of further research. 

This paper considers one particular type of Bayesian approach, i.e., Gaussian processes \cite{RasmussenW06}. The method has its roots in Stochastic processes—a field of study dedicated to modeling random processes with Probability theory \cite{Klebaner12,Rosenthal06}. Most problems of interest are usually not deterministic processes, or even if they are, one might not have access to all the information needed to model it as such. Stochastic processes mathematically accommodate such uncertainties, and Gaussian processes are one particular variant of Stochastic processes. 

I start my exposition by detailing Gaussian processes, their advantages, and their disadvantages. However, the main focus of this survey is Deep Gaussian Processes (DGPs). I will describe some of the Gaussian Processes' prominent variants crucial for building DGPs and explain the critical DGP approaches. 

\section{Preliminary}

A collection of random variables is called a Gaussian Process(GP) if the joint distribution of any finite subset of its variables is a Gaussian. Gaussian processes can be thought of as a reinterpretation or a generalization of Gaussian distributions. Gaussian processes are essentially Gaussian distributions with infinite dimensions. Each dimension of the distribution corresponds to an input data sample, and the distribution mean represents each data input's associated label. 

Gaussian processes are fascinating because of their analytical properties. Gaussian processes being Gaussian distributions, have closed-form analytical solutions to summation, conditioning, and marginalization under certain conditions \cite{Murphy12}. Given a Gaussian distribution on a dataset $\mathbf{X}$ with mean $\boldsymbol{\mu}$ and covariance $\mathbf{\Sigma}$, the marginals and conditionals are given by the following

$$
\mathbf{X}=\left(\begin{array}{l}
\mathbf{X}_{1} \\
\mathbf{X}_{2}
\end{array}\right), \quad \boldsymbol{\mu}=\left(\begin{array}{l}
\boldsymbol{\mu}_{1} \\
\boldsymbol{\mu}_{2}
\end{array}\right), \quad \boldsymbol{\Sigma}=\left(\begin{array}{ll}
\boldsymbol{\Sigma}_{11} & \boldsymbol{\Sigma}_{12} \\
\boldsymbol{\Sigma}_{21} & \boldsymbol{\Sigma}_{22}
\end{array}\right)
$$
$$
\begin{aligned}
p\left(\mathbf{X}_{1}\right) &=\mathcal{N}\left(\mathbf{X}_{1} \mid \boldsymbol{\mu}_{1}, \boldsymbol{\Sigma}_{11}\right) \\
p\left(\mathbf{X}_{2}\right) &=\mathcal{N}\left(\mathbf{X}_{2} \mid \boldsymbol{\mu}_{2}, \mathbf{\Sigma}_{22}\right) \\ \\
p\left(\mathbf{X}_{1} \mid \mathbf{X}_{2}\right) &=\mathcal{N}\left(\mathbf{X}_{1} \mid \boldsymbol{\mu}_{1 \mid 2}, \boldsymbol{\Sigma}_{1 \mid 2}\right) \\
\boldsymbol{\mu}_{1 \mid 2} &=\boldsymbol{\mu}_{1}+\boldsymbol{\Sigma}_{12} \boldsymbol{\Sigma}_{22}^{-1}\left(\mathbf{X}_{2}-\boldsymbol{\mu}_{2}\right) \\
\boldsymbol{\Sigma}_{1 \mid 2} &=\boldsymbol{\Sigma}_{11}-\boldsymbol{\Sigma}_{12} \boldsymbol{\Sigma}_{22}^{-1} \boldsymbol{\Sigma}_{21}
\end{aligned}
$$
Moreover, as mentioned before, deep neural networks require the specification of model architectures which are usually application-specific.  However, GPs do not suffer from such issues as they are non-parametric. GPs consider the entire function space and marginalize it to obtain their predictions. The approach has two main advantages; first, one need not decide the model's complexity. The second, the marginalization process, induces Occum's Razor \cite{RasmussenW06}, thereby overcoming any overfitting issues. 
 
One might wonder if modeling any arbitrary process as a Gaussian would result in poor performance. However, Gaussians are omnipresent, and the Central Limit Theorem applies to them  \cite{RasmussenW06}. The theorem states that any dataset's mean follows a Gaussian distribution as the number of samples from the distribution increases. So, assuming enough data is available, usually, a Gaussian approximation results in reasonable estimates. 

Another attractive property of Gaussians is that the distribution has the maximum entropy given a dataset's mean and variance \cite{Murphy12}. The first two moments, mean and variance, are usually the only metrics that we can accurately compute from data. As a result, using Gaussians to estimates distributions results in the most informative data distribution estimates. 

Finally, another essential property of Gaussian processes was established by Neil \cite{Neal96} where he showed that GPs are equivalent to a single-layered Perceptron with infinite hidden units. Perceptrons were, in turn, known to be universal approximators that can approximate any function given enough data \cite{HornikSW89}.  

\section{Gaussian Processes}
I have detailed the critical advantages of Bayesian approaches and why researchers are interested in Gaussian processes specifically. This section further elaborates on GPs. I give an intuition of GPs; their mathematical formulation \cite{RasmussenW06, Murphy12}, and an intuitive interpretation of the terms in its formulation.  Furthermore, I will explain Kernel functions and list some limitations of GPs. 

\begin{figure}[htp]
    \centering
    \includegraphics[width=\linewidth]{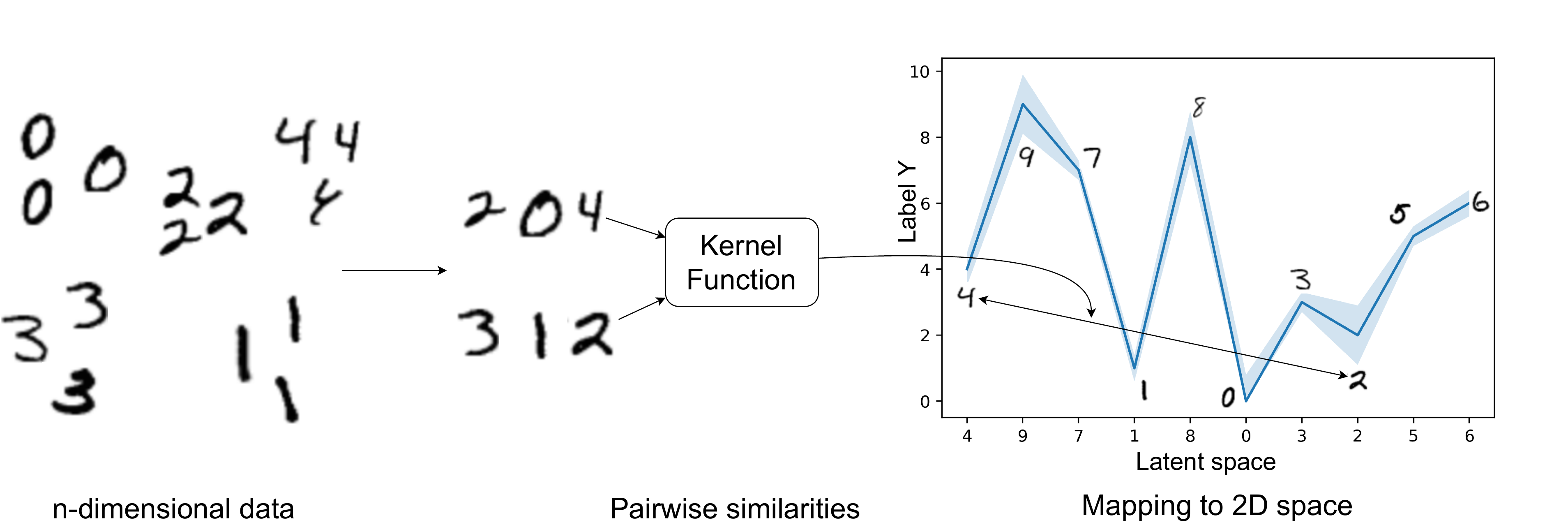}
    \caption{Illustration of a Gaussian Process on the MNIST dataset.}
    \label{fig:mnist_gp}
\end{figure}

GPs, as mentioned before, are Gaussian distributions interpreted as probabilistic mappings from an input data space to its output label space. This mapping has a convenient interpretation in function space. Formally, we assume a function $f$, maps an input space $\mathbf{X}$ to an output space $\mathbf{y}$. This mapping is characterized by a noise term $\epsilon$.
$$
\mathbf{y}=f(\mathbf{X})+\boldsymbol{\epsilon}, \boldsymbol{\epsilon} \sim \mathcal{N}\left(\mathbf{0}, \beta^{-1} \mathbf{I}\right)
$$
In a non-probabilistic approach, a parametric form for the function $f$ is considered, and Maximum A Posteriori (MAP) is used to estimate the function parameters. GPs, on the other hand, assume a distribution over the functions $f$. The function values are also called latent variables, and any predictions in GPs involve marginalizing over the entire function space. Fig.~\ref{fig:mnist_gp} illustrates this process. Furthermore, the distribution over the functions is parametrized by a mean vector and a covariance matrix. The covariance is computed with a similarity metric known as a kernel function $K$.  
$$
\begin{aligned}\mathbf{f}|\mathbf{X} & \sim \mathcal{G} \mathcal{P}\left(\mu(\mathbf{X}), K\left(\mathbf{X}, \mathbf{X}\right)\right) \\
\mu(\mathbf{X}) &=\mathbb{E}[\mathbf{f}(\mathbf{X})] \\
K\left(\mathbf{X}, \mathbf{X}\right) &=\mathbb{E}\left[(\mathbf{f}(\mathbf{X})-\mu(\mathbf{X}))\left(\mathbf{f}\left(\mathbf{X}\right)-\mu\left(\mathbf{X}\right)\right)\right]
\end{aligned}
$$
Moreover, by treating the function outputs as noisy versions of the labels, the distribution over $\mathbf{y}$ conditioned on the input $\mathbf{X}$ can be achieved as follows. 
$$
\begin{aligned}
p(\mathbf{y} \mid \mathbf{f}) &= \mathcal{N}\left(\mathbf{f}, \beta^{-1} \mathbf{I}\right) \\
p(\mathbf{y} \mid \mathbf{X}) &= \int p(\mathbf{y} \mid \mathbf{f})p(\mathbf{f} \mid \mathbf{X}) d \mathbf{f}
\end{aligned}
$$
The above marginalization is the likelihood of the dataset. It is used to fit hyperparameters involved in the GP formulation by maximizing the training dataset's likelihood. However, we are also interested in predicting labels for new data samples. We can compute the predictions by considering a joint distribution over the function values of data  $\mathbf{X}$ and $\mathbf{X}_*$ from the training and testing datasets denoted by  $\mathbf{f}$ and $\mathbf{f}_*$ respectively. 
$$
\left[\begin{array}{l}
\mathbf{f} \\
\mathbf{f}_{*}
\end{array}\right] \sim \mathcal{N}\left(\mathbf{0},\left[\begin{array}{ll}
K(\mathbf{X}, \mathbf{X}) & K\left(\mathbf{X}, \mathbf{X}_{*}\right) \\
K\left(\mathbf{X}_{*}, \mathbf{X}\right) & K\left(\mathbf{X}_{*}, \mathbf{X}_{*}\right)
\end{array}\right]\right)
$$
We are interested in $\mathbf{f}_*$ as it is the noise-free realization of $\mathbf{y}_*$. The posterior distribution of $\mathbf{f}_*$ can be obtained by conditioning it with $\mathbf{X}_*$, $\mathbf{X}$, and $\mathbf{y}$ as follows. 
$$
p(\mathbf{f_*} \mid \mathbf{X}_*, \mathbf{X}, \mathbf{y}) = \frac{1}{p({\mathbf{y}})} \int p(\mathbf{y} \mid \mathbf{f}) p(\mathbf{f}, \mathbf{f_*}) d \mathbf{f}
$$
$$
\begin{aligned}
p(\mathbf{f_*} \mid \mathbf{X}_*, \mathbf{X}, \mathbf{y}) \sim \mathcal{N}(& K\left(\mathbf{X}_{*}, \mathbf{X}\right) K(\mathbf{X}, \mathbf{X})^{-1} \mathbf{f}, \\
&\left.K\left(\mathbf{X}_{*}, \mathbf{X}_{*}\right)-K\left(\mathbf{X}_{*}, \mathbf{X}\right) K(\mathbf{X}, \mathbf{X})^{-1} K\left(\mathbf{X}, \mathbf{X}_{*}\right)\right)
\end{aligned}
$$
The above conditional prediction distribution's mean is treated as the model prediction since a Gaussian distribution's mode coincides with its mean. The mean term of the predictive distribution can be interpreted as a weighted average of the training set labels. The weights are defined by the kernel matrices, which assign higher weights to data points close to the test samples in feature space. 

The variance term has two parts; the first term on the left can be assumed to be the prior variance from the testing data. And the second term can be thought of as the variance induced by the training data. By subtracting the two terms, the initial test data variance is reduced by the training data's information. 

\subsection{Kernel Function}
Unlike a conventional Gaussian distribution, a GP has infinite dimensions; therefore, explicitly defining a covariance matrix would be infeasible. GPs resort to using kernel functions to determine the covariance matrix, allowing for a compact representation of the kernel matrix. 

Kernel functions measure data similarity, and GPs use it to produce a positive semi-definite (PSD) matrix. Kernel functions can be interpreted as dot products between data points in a high dimensional manifold without explicitly mapping them to such a manifold. Hence, these functions allow us to define meaningful similarity metrics relevant to a given task in a computationally efficient manner.  

Numerous kernel functions have been used in literature, each with its own set of properties that make it useful for different tasks. Enumerating and explaining all such functions is out of this paper's scope, and the readers are referred to \cite{RasmussenW06} for a detailed overview. However, two kernels are often used in literature, particularly the Squared Exponential kernel and the Automatic Relevance Detection (ARD) kernel functions detailed below. 

The Squared Exponential kernel, also known as the Radial Basis Function and Gaussian kernel, has an exponential form. It has two parameters, the length scale $\ell$, which determines how close a pair of data points should be in order to be correlated. And it has a variance scale $\sigma$, which quantifies the variance induced from data points. The function is parameter efficient as it has only two parameters, and its exponential form allows for analytical computations of integrals in some cases.  For a pair of inputs $\mathbf{x}$, $\mathbf{x}^{'} \in \mathbb{R}^q$, the Gaussian kernel is defined as follows
$$
K\left(\mathbf{x}, \mathbf{x}^{'}\right)=\sigma^{2} \exp \left(-\frac{1}{2 \ell^{2}} \sum_{j=1}^{q}\left(\mathbf{x}_{j}-\mathbf{x}^{'}_{j}\right)^{2}\right)
$$
Another important kernel function is the Automatic Relevance Detection (ARD) kernel. It has a similar form to the Gaussian kernel. However, it has weight parameters for each dimension of the input space. We can use the weights to scale down irrelevant features and increase the importance of the relevant ones. 
$$
K\left(\mathbf{x}, \mathbf{x}^{'}\right)=\sigma^{2} \exp \left(-\frac{1}{2} \sum_{j=1}^{q}w_j\left(\mathbf{x}_{j}-\mathbf{x}^{'}_{j}\right)^{2}\right)
$$
The weight parameter in the ARD kernel function can be defined manually. However, such an approach is infeasible as we usually don't know in advance if a feature is essential or not.  In practice, the weights are determined from the training dataset, which allows us to obtain kernel function parametrizations that are most suited for the task. Indeed, there are two primary approaches to determine the function parameters, Maximum A Posteriori (MAP) and the Bayesian approach. 

The MAP approach considers the log probability of the dataset $\log p(\mathbf{y} \mid \mathbf{X})$. The optimal parameters are computed by taking the log probability derivative with respect to the kernel parameters. Depending on the dataset size, the parameters can be calculated analytically but, in some cases, one needs to resort to approaches like gradient descent to determine the parameters. 

$$
\log p(\mathbf{f} \mid \mathbf{X})=-\frac{1}{2} \mathbf{f}^{\top} K^{-1} \mathbf{f}-\frac{1}{2} \log |K|-\frac{n}{2} \log 2 \pi \\
$$
$$
\log p(\mathbf{y} \mid \mathbf{X})=-\frac{1}{2} \mathbf{y}^{\top}\left(K+\sigma_{n}^{2} I\right)^{-1} \mathbf{y}-\frac{1}{2} \log \left|K+\sigma_{n}^{2} I\right|-\frac{n}{2} \log 2 \pi
$$
A limitation of using MAP to determine the kernel parameters is that it could lead to overfitting. The problem's severity varies depending on the dataset size, the GP's variant, and the kernel function being used. Nonetheless, we can mitigate the problem by using a Bayesian approach. 

A Bayesian approach defines a distribution over the kernel parameters and computes the posterior parameter distribution conditioned on the training dataset. Although the Bayesian approach is ideal, it is usually intractable to analytical computations. As such, one has to resort to sampling approaches such as Hamiltonian Monte Carlo (HMC) methods \cite{Bishop06} to determine the parameter distributions. However, HMC methods have their limitations and introduce additional computational costs that might be infeasible. 

\subsection{Limitations}

Although GPs have several advantages, they also have a few key limitations which hinder their use in most machine learning problems. Particularly, there are three main problems:

\begin{enumerate}
\item Computation cost
\item Storage cost
\item Hierarchical feature extraction
\end{enumerate}

\begin{figure}[htp]
    \centering
    \includegraphics[width=0.6\linewidth]{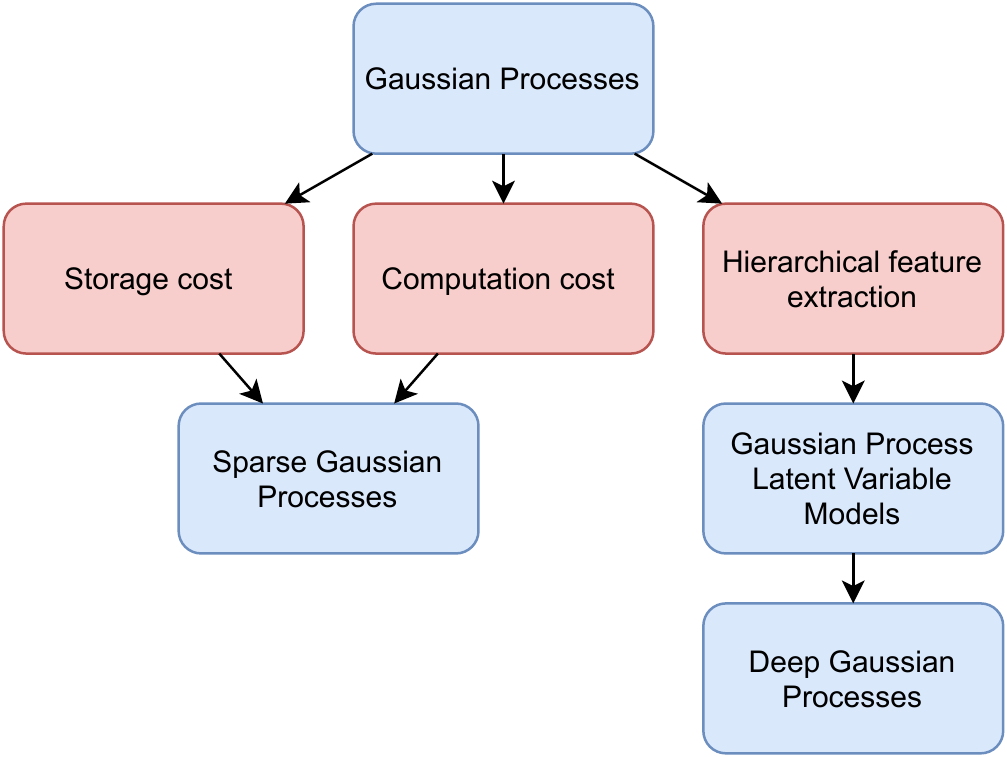}
    \caption{Overview of Gaussian Processes and their variants.}
    \label{fig:overview}
\end{figure}

The computational cost of GPs can be rather substantial; one needs to invert the kernel matrix to get the predictive distribution of a GP. The kernel matrix's size is $n \times n$ where $n$ is the number of data points in the training dataset. Inverting such a matrix takes $O(n^3)$ computation time. Furthermore, once the kernel matrix inverse is available, it takes $O(n)$ and $O(n^2)$ time to determine the predictive distribution's mean and variance of a new data point.

Additionally, since GPs require the entire training dataset's storage, the storage cost is $O(n^2)$. Depending on the size of the dataset, the storage costs substantially limit the scalability of the approach. Moreover, if the GP were to be used in an environment where the training dataset size keeps increasing, the computation and storage costs could overwhelm the entire process, rendering the benefits of GP far too expensive. As such, GPs are usually only feasible for datasets with about $1000-3000$ data points.

Another major drawback of GPs is the lack of kernel functions capable of handling structured data wherein one needs to consider hierarchical feature extraction to determine the similarity of a pair of data points properly. Such an issue often arises in data such as images but, it is also prevalent in simpler vector datasets. Conventional kernel functions are ill-equipped to handle such correlations, and one needs to resort to deep feature extraction like the ones used in Deep learning models. However, such feature extraction still needs to be confined to a Bayesian framework to retain a GP's advantages.

Sparse Gaussian Processes address the computational and storage costs. And the feature extraction issue is addressed by Deep Gaussian Processes. I explain some of the prominent methods for Sparse and Deep GPs that have been developed over the last two decades in the following sections. Fig:~\ref{fig:overview} shows a flow chart that relates the limitations to the GP varients that address them.  

\section{Sparse Gaussian Processes}

\begin{wrapfigure}{r}{0.5\textwidth}
  \begin{center}
    \includegraphics[width=0.2\textwidth]{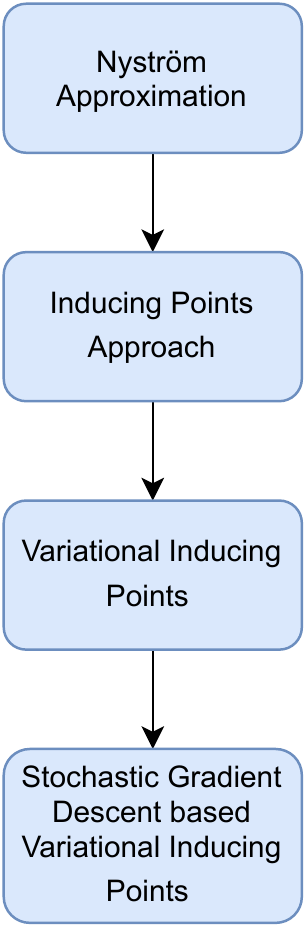}
  \end{center}
    \caption{Overview of Sparse Gaussian Processes.}
    \label{fig:sgp}
\end{wrapfigure}

Given the computational and storage requirements that hinder the widespread use of GPs, a substantial number of papers have tried to address the problem and are collectively referred to as Sparse Gaussian Processes (SGPs), Fig.~\ref{fig:sgp} depicts the prominent methods coverd in this section. The terminology stems from the way most of these approaches address the issue. Because the primary problem is the inversion of a covariance matrix, most methods try to introduce sparsity and reduce the matrix size that needs to be inverted while retaining the original matrix's performance. 

This section focuses on some of the well-known methods crucial for developing some of the Deep Gaussian Process methods that will be detailed in the upcoming section. A complete overview of all SGPs is outside this survey's scope; the readers are referred to \cite{LiuOSC20} for a thorough summary. 

The Nyström approximation of \cite{WilliamsS01} is a well know approach to reducing the inversion cost of the covariance matrix in GPs. The Nyström approximation allows one to generate a low rank approximation of any kernel matrix. The approach is applied to GPs by selecting $m$ data points with $m<<n$ from the training set.  Then a low rank approximation $\tilde{K}$  to the kernel matrix is computed as shown below

$$\tilde{K} = K_{n,m} K_{m,m}^{-1}  K_{m,n}$$
Here, $K_{n, m}$ represents the kernel matrix computed from the $n$ and $m$ data points in the training dataset and the selected subset, respectively. The same notation is used for the other kernel matrices. The approximation needs the inversion of only an $m \times m$ matrix, thereby reducing the computation cost from $ O(n^3)$ to  $O(m^3)$.

However, the approximation assumes that the data comes from a low rank manifold, which would be the case if the data dimension $d<n$. In such a case, the low rank approximation would be exact and result in no information loss.  But, the selected $m$ data points also influence the approximation. It is possible to have data points that result in a poor approximation even if the data is from a low dimensional manifold. 

In practice, most datasets have more data points than the number of features; thus, the method is applicable in most cases. However, selecting the data points is crucial to the approximation's performance. Williams and Seeger used a random subset of $m$ data points in their approach  \cite{WilliamsS01}. Although the method works, the unsophisticated data selection process limited the method's performance. 

Snelson and Ghahramani \cite{SnelsonEZ06} addressed the subset selection for Nyström approximation by treating the subset as model parameters and termed them pseudo data. The pseudo data is assumed to be synthetic and does not necessarily correspond to any data point available in the training dataset. Indeed, they could take values that are some combination of the training dataset. 

The pseudo points were computed using maximum likelihood. However, to use maximum likelihood, one needs to parameterize the GP with pseudo points appropriately. Snelson and Ghahramani \cite{SnelsonEZ06} introduced a distribution over the pseudo points and considered a joint distribution over latent representations of data from the training, testing and, pseudo data points given by $\mathbf{f}, \mathbf{f_*}$ and $\mathbf{m}$. The authors then marginalized the pseudo points to obtain the posterior distribution as shown below
$$
\begin{aligned}
p(\mathbf{y} \mid \mathbf{X}, \mathbf{m}) &= \mathcal{N}\left(\mathbf{y} \mid K_{nm} K_{mm}^{-1} \mathbf{m}, \mathbf{\Lambda}+\sigma^{2} \mathbf{I}\right) \\
\mathbf{\Lambda} &=\operatorname{diag}(\boldsymbol{\lambda}) \\
\lambda_{n} &= K_{nn}-K_{mn}^{\top} K_{mm}^{-1} K_{mn} \\
p\left(\mathbf{f}_{*}, \mathbf{f}\right) &= \int p\left(\mathbf{f}_{*}, \mathbf{f}, \mathbf{m}\right) \mathrm{d} \mathbf{m} =\int p\left(\mathbf{f}_{*}, \mathbf{f} \mid \mathbf{m}\right) p(\mathbf{m}) \mathrm{d} \mathbf{m} \\
p(\mathbf{m}) &=\mathcal{N}\left(\mathbf{0}, K_{\mathbf{m}, \mathbf{m}}\right)
\end{aligned}
$$
Although using maximum likelihood to determine the pseudo points' distribution works in practice, using maximum likelihood runs the risk of overfitting. It would be ideal to use a Bayesian approach to compute the pseudo points distribution conditioned with the training set. Unfortunately, such an approach is infeasible as it becomes intractable to analytical solutions of the pseudo points. Moreover, the approach works by assuming that the joint distribution $p\left(\mathbf{f}_{*}, \mathbf{f}\right)$ can be partitioned as follows

$$
p\left(\mathbf{f}_{*}, \mathbf{f}\right) =\int p\left(\mathbf{f}_{*}, \mid \mathbf{m}\right)p\left(\mathbf{f} \mid \mathbf{m}\right) p(\mathbf{m}) \mathrm{d} \mathbf{m} \\
$$

The assumption limits the information a GP obtains from the training set to be induced only through the pseudo set. Hence, the pseudo points are also referred to as inducing inputs. The factorization assumption limits the capacity of the model and affects the accuracy of the model. Notably, the prior distribution that is assumed for the pseudo set substantially influences the results. 

Snelson and Ghahramani \cite{SnelsonEZ06} treated the pseudo points as hyperparameters and introduced a prior over them, resulting in an inaccurate posterior compared to a vanilla GP. The inaccuracy was a consequence of the formulation of the kernel approximation. Titsias \cite{Titsias09} addressed the outfitting and inexact posterior issue by considering a variational approach. The approach introduces a lower bound that can be optimized to determine the inducing inputs and the kernel hyperparameters. The bound shown below, can be used to solve for the inducing points and kernel hyperparameters. We can then use the inducing points for computing the predictive distribution.

$$
\begin{aligned}
\mathcal{F} &= \log \left[\mathcal{N}\left(\mathbf{y} \mid \mathbf{0}, \sigma^{2} I+Q_{nn}\right)\right]-\frac{1}{2 \sigma^{2}} \operatorname{Tr}(\widetilde{K}) \\
Q_{nn} &= K_{nm} K^{-1}_{mm} K_{mn} \\
\widetilde{K} &=  K_{nn}-K_{nm} K^{-1}_{mm} K_{mn} 
\end{aligned}
$$
However, the marginal in \cite{Titsias09} didn't have the factorization required to apply stochastic gradient descent. Hensman et al. \cite{HensmanFL13} improved on the work of Titsias \cite{Titsias09} by developing a new bound that could be optimized with stochastic gradient descent. Unlike Titsias's approach, the method in \cite{HensmanFL13} did not require the entire dataset at once to compute the variational parameters. It used the following bound that could be optimized with stochastic gradient descent. 
$$
\begin{aligned}
\mathcal{F}=\sum_{i=1}^{n}\{& \log \mathcal{N}\left(\mathbf{y}_{i} \mid K_{i}^{\top} K_{mm}^{-1} \mathbf{v}, \beta^{-1}\right) \\
&\left.-\frac{1}{2} \beta \widetilde{k}_{i, i}-\frac{1}{2} \operatorname{tr}\left(\mathbf{S} \mathbf{\Lambda}_{i}\right)\right\} \\
&-\mathrm{KL}(q(\mathbf{u}) \| p(\mathbf{u})) \\
\end{aligned}
$$

$$
\begin{aligned}
q(\mathbf{u}) &= \mathcal{N} (\mathbf{u} \mid \mathbf{v}, \mathbf{S}) \\
\mathbf{\Lambda} &= \beta K^{-1}_{mm} k_i k_i^{\top} K^{-1}_{mm}
\end{aligned}
$$
Here, $\mathbf{u}$ is the set of feature space representations of the inducing points, an0d $k_i$ is the $i^{th}$ column of $K_{mn}$.  Hensman et al. showed the approach to scale well to large datasets while retaining the reduced model complexity of  $O(m^3)$. 

\section{Gaussian Process Latent Variable Model}
The methods discussed so far primarily addressed the computation and storage cost issues. This section introduces a variant of GPs that can be trained in an unsupervised manner. Gaussian Process Latent Variable Models (GPLVMs)  \cite{Lawrence04,Lawrence05} assume the feature space is a latent space with unknown data distribution. The latent space distribution in then learned during the training phase. Although the method seems irrelevant, it plays a significant role in some of the Deep Gaussian Processes that I will cover in the next section.  

Lawrence \cite{Lawrence04,Lawrence05} showed that if the function space of a GP is constrained to linear function space, the GP can be interpreted as a probabilistic variant of Principal Component Analysis (PCA). Moreover, if the function space is relaxed to a non-linear space defined by a kernel function, it can be interpreted as probabilistic non-linear PCA. The approach assumes a standard Gaussian prior over the input space and maximizes the log probability of the dataset $p( \mathbf{y} \mid \mathbf{X}, \beta)$ with respect to the input data likelihood $\mathbf{X}$. 

The input data or latent space distribution can not be computed analytically because of the non-linearities introduced by the kernel function. However, Lawrence \cite{Lawrence04, Lawrence05} showed that the distribution could be estimated using the Expectation-Maximization algorithm. But, the approach returns only the mode of the data distribution as a consequence.

Additionally, GPLVMs were shown to be useful for reconstructing inputs with partially observed features. Such a scenario often occurs in image reconstruction or denoising tasks.

Although GPLVMs are very useful for unsupervised tasks, the original method assumed access to the full kernel matrix, which entails storing the entire training dataset and inverting an $n \times n$ kernel matrix. Lawrence \cite{Lawrence07} addressed this issue by showing that most sparse Gaussian process approaches can be turned into GPLVMs. However, the approach still gave a MAP estimate of the latent space and risked overfitting to the training dataset. 

Titsias and Lawrence \cite{TitsiasL10} addressed the overfitting issue by proposing a Bayesian approach. Instead of finding a MAP solution to the latent space, they proposed a variational approach. However, naive use of the variational approach to finding the data distribution introduces intractabilities. Titsias and Lawrence addressed the problem by incorporating Titsias's variational approach to SGPs \cite{Titsias09}. The introduction of pseudo points \cite{Titsias09} canceled out the intractable terms in the variational bound for GPLVMs and resulted in a feasible optimization bound shown below 

$$
\begin{aligned}
\mathcal{F}(q) \geq \log \left(\int e^{\left\langle\log \mathcal{N}\left(\mathbf{y}_{d} \mid \boldsymbol{\alpha}_{d}, \beta^{-1} I\right)\right\rangle_{q(\mathbf{X})}}p\left(\mathbf{m}_{d}\right) d \mathbf{m}_{d}\right) \\
-\frac{\beta}{2} \operatorname{Tr}\left(\left\langle K_{nn}\right\rangle_{q(\mathbf{X})}\right)+\frac{\beta}{2} \operatorname{Tr}\left(K_{mm}^{-1}\left\langle K_{mn} K_{nm}\right\rangle_{q(\mathbf{X})}\right)
\end{aligned}
$$
Here, $q$ is the variational distribution over pseudo points $\mathbf{m}$ and $\mathbf{\alpha} = K_{nm}K^{-1}_{mm}\mathbf{m}$. The subscript $d$ is used to denote each dimension of the features.  

GPLVMs have been applied to numerous applications and their variants that we did not cover here. The readers are referred to \cite{LiC16} for an in-depth survey of GPLVMs. 

\section{Deep Gaussian Processes}
Although SGPs addressed the computation cost issue, GPs remain inapplicable to a lot of applications. The reason being the kernel function. The most commonly used kernel functions have relatively simple similarity metrics. However, in specific datasets, one might have to use a different similarity metric in different regions of the input space. A similarity metric that can extract such features would have to utilize a hierarchical structure for feature extraction. 

One strategy to the problem would be to stack GPs similar to how Perceptrons are stacked in an MLP. But, stacking GPs such that one layer's output becomes the following layer's input makes them highly non-linear and intractable to analytical solutions. Moreover, a stacked GP would not even correspond to a GP anymore as the posterior distribution can take on any arbitrary distribution. However, such methods are usually referred to as Deep Gaussian Processes (DGPs). Several authors have tried to model and fit such models; this section explains such methods' development. 

One of the earliest DGP approaches is that of Lawrence and Moore \cite{LawrenceM07}. They considered a GPLVM model, but a GP was assumed for the input space prior distribution, making it a two layered DGP. The DGP resulted in the following likelihood function, which cannot be marginalized analytically. 

$$
p\left(\mathbf{y}\mid\mathbf{t}\right) =\int p\left(\mathbf{y}\mid\mathbf{X}\right) p\left(\mathbf{X}\mid\mathbf{t}\right)\mathrm{d} \mathbf{X} \\
$$

Here, $\mathbf{t}$ is the input GP at the input layer, and $\mathbf{X}$ are the intermediate representations passed to the second layer GP. Lawrence and Moore considered a MAP solution to the above problem. This was achieved by maximizing the following. 
$$
p\left(\mathbf{X}\mid\mathbf{y, t}\right) =\log p\left(\mathbf{y}\mid\mathbf{X}\right) + \log p\left(\mathbf{X}\mid\mathbf{t}\right) + \text {const} \\
$$

The authors also showed that deeper hierarchies could be modeled with such an approach. However, the model was limited to a MAP solution which is highly susceptible to overfitting.  

Damianou et al. \cite{DamianouTL11, damianou15} proposed a variational approach to the overfitting problem. They also considered a 2-layered stacked GP, but a naive variational bound for such a model introduces intractabilities similar to a GPLVM.  However, the authors showed that the variational approach used for Bayesian GPLMVs in \cite{TitsiasL10} could also be utilized to formulate a variational bound for a 2-layered GP. The final bound is shown below, with $q(\mathbf{X})$ as the variational distribution

$$
\mathcal{F}=\hat{\mathcal{F}}-\operatorname{KL}(q(X) \| p(X \mid \mathbf{t})) \\
\hat{\mathcal{F}}=\int q(X) \log p(y \mid \mathbf{f}) p(\mathbf{f} \mid X) \mathrm{d} X \mathrm{~d} \mathbf{f}
$$

Furthermore, Damianou and Lawrence \cite{DamianouL13} improved the bound shown above by generalizing the variational bound to a DGP with an arbitrary number of layers. The bound shown below can be used on a DGP with two or more layers. 

$$
\begin{aligned}
\mathcal{F}&=\mathbf{g}_{Y}+\mathbf{r}_{X}+\mathcal{H}_{q(\mathbf{X})}-\operatorname{KL}(q(\mathbf{Z}) \| p(\mathbf{Z})) \\
\mathbf{g}_{Y}&=g\left(\mathbf{Y}, \mathbf{F}^{Y}, \mathbf{U}^{Y}, \mathbf{X}\right) \\
\quad&=\left\langle\log p\left(\mathbf{Y} \mid \mathbf{F}^{Y}\right)+\log \frac{p\left(\mathbf{U}^{Y}\right)}{q\left(\mathbf{U}^{Y}\right)}\right\rangle_{p\left(\mathbf{F}^{Y} \mid \mathbf{U}^{Y}, \mathbf{X}\right) q\left(\mathbf{U}^{Y}\right) q(\mathbf{X})} \\
\mathbf{r}_{X}&=r\left(\mathbf{X}, \mathbf{F}^{X}, \mathbf{U}^{X}, \mathbf{Z}\right) \\
&=\left\langle\log p\left(\mathbf{X} \mid \mathbf{F}^{X}\right)+\log \frac{p\left(\mathbf{U}^{X}\right)}{q\left(\mathbf{U}^{X}\right)}\right\rangle_{p\left(\mathbf{F}^{X} \mid \mathbf{U}^{X}, \mathbf{Z}\right) q\left(\mathbf{U}^{X}\right) q(\mathbf{X}) q(\mathbf{Z})}
\end{aligned}
$$

\begin{figure}[htp]
    \centering
    \includegraphics[width=0.45\textwidth]{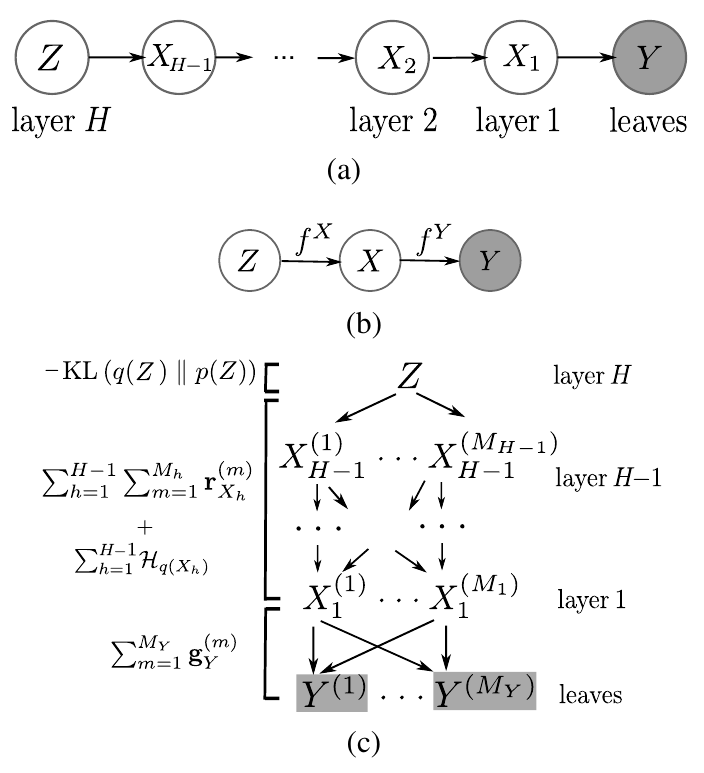}
    \caption{Deep Gaussian Process model overview \cite{DamianouL13}.}
    \label{fig:dgp}
\end{figure}

Here, $\mathbf{Y}$ represents the output multidimensional label space, $\mathbf{Z}$ represents the latent variables in the input layer, and $\mathbf{X}$ represents the latent inputs in intermediate layers. $\mathbf{U}$ and $\mathbf{F}$ are the values of the latent function corresponding to inducing points and latent inputs, respectively; their superscript denotes the layer to which they belong. Additionally, $\mathcal{H}$ represents the entropy of the distribution shown in its subscript, and $\operatorname{KL}$ is the standard KL-divergence. Fig.~\ref{fig:dgp} shows the DGP model architecture from \cite{DamianouL13}. 

Again, the crux of the approach relied on the variational trick of introducing inducing points as presented in \cite{TitsiasL10}. Damianou and Lawrence conducted experiments on the MNIST dataset wherein they showed a 5-layer DGP could be used for the image classification task. 

A limitation of the approach in \cite{DamianouL13} is that the number of variational parameters that need to be learned increases linearly with the number of data points in the training set. And it involved inverting matrices which is a computationally expensive operation, thereby limiting its scalability. Dai et al. \cite{DaiDGL15} addressed this by introducing a back-constraint. The constraint allowed them to define the latent variables' mean terms as deterministic functions of the latent variables themselves via an MLP. The approach reduced the number of variational parameters. Moreover, Dai et al. \cite{DaiDGL15} also showed that their approach could be trained in a distributed manner, allowing the model to be scaled to large datasets.

Salimbeni and Deisenroth \cite{SalimbeniD17} recently proposed an approach that addressed the layer independence issue of prior DGP approaches. The DGP in \cite{DamianouL13} assumed independence of GPs across layers and only considered the correlations within a layer. However, Salimbeni and Deisenroth argue that such an approach is equivalent to a single GP with each GP's input coming from a GP itself. The authors also stated that they found that some layers get turned off when using the DGP in \cite{DamianouL13}. 

Salimbeni and Deisenroth \cite{SalimbeniD17} presented a new variational bound that retains the exact model posterior similar to \cite{DamianouL13} while maintaining the correlations both within and across adjacent layers. However, Salimbeni and Deisenroth showed such an approach is infeasible to analytic computations, but the bound could still be optimized with MCMC sampling techniques. Such an approach is computationally expensive. But, it can be parallelized by exploiting the factorization of the DGP across output dimensions. Furthermore, the method also required sampling methods during inference, but its performance is substantially better than prior works.  

$$
\mathcal{F}=\sum_{i=1}^{N} \mathbb{E}_{q\left(\mathbf{f}_{i}^{L}\right)}\left[\log p\left(\mathbf{y}_{n} \mid \mathbf{f}_{n}^{L}\right)\right]-\sum_{l=1}^{L} \mathrm{KL}\left[q\left(\mathbf{U}^{l}\right) \| p\left(\mathbf{U}^{l} ; \mathbf{Z}^{l-1}\right)\right]
$$

In the optimization bound shown above from \cite{SalimbeniD17}, the subscripts are used to denote each data sample in the dataset, and superscripts are used to indicate a layer in the DGP. The rest of the terms follow the same convention as the one used by Damianou et al. \cite{DamianouL13}.

I briefly mentioned that DGPs do not necessarily correspond to Gaussian processes. Still, the methods discussed so far do model the posterior distribution as a Gaussian, each with its assumptions. Havasi et al. \cite{HavasiLF18} presented a technique that departs even more from a conventional GP. The authors show that since a Gaussian is uni-modal, using it to model the posterior will result in poor results. Instead, they suggest using a multi-modal distribution that can better capture the true posterior distribution.  

However, it is not possible to formulate an analytical solution to a multi-modal posterior. We can use variational inference to learn a multi-modal posterior. Still, one needs to determine the exact form of the variational distribution, which is difficult as we usually don't know the posterior distribution in advance. Havasi et al. \cite{HavasiLF18} circumvent this issue by using Stochastic Gradient Hamiltonian Monte Carlo (SGMCMC) \cite{ChenFG14} method to estimate the posterior. The approach can determine the inducing points by sampling from the true posterior instead of using a variational distribution.  

Although the approach far exceeds the performance of prior DGPs and is the current state-of-the-art, it still has its limitations. Remarkably, the SGMCMC method is difficult to tune as it introduces its own parameters in addition to the ones that already have to be estimated for DGP. Several MCMC method variants attempt to improve upon SGMCMC, but none of those approaches have been applied to DGPs. 

The DGPs we discussed so far attempted to develop variants of GPs that can model hierarchical features in data, which was done by assuming a feed forward network with each node of the network being modeled as a GP. It is the most popular method to address the problem and has resulted in approaches that can get reasonably promising results. However, there are other approaches as well which do not consider such an explicit feed forward network.

Wilson et al. \cite{WilsonHSX16} proposed a method that uses deep neural networks as the kernel function, termed deep kernel. Unlike a Gaussian kernel, the deep kernel produced a vector output, and a GP was assigned to each of the vector elements. Wilson et al. further combined the GPs with an additive structure that facilitated its training with an analytical bound. 

Wilson et al. \cite{WilsonHSX16} showed that their approach was good at several tasks. However, the deep neural network architecture needs to be task specific, and its parameters are susceptible to overfitting given its large parameter count.  

Another interesting perspective on DGPs was presented by Lee et al. \cite{LeeBNSPD18}. Until now, all the discussed GPs with linear latent functions are combined in different ways to achieve an aggregate non-linear latent function space. Lee et al. developed a method that considers an entire function space that consists of non-linear functions. Unlike prior approaches, the function space was not constrained to a specific subspace as a consequence of a particular kernel function being used. The approach can be regarded as a generalization of Neil \cite{Neal96} who showed the equivalence of an infinitely wide single layer neural network to a GP. Lee et al. showed the equivalence of a GP to an infinitely wide deep neural. 

Lee et al. \cite{LeeBNSPD18} showed the approach was on par with some neural networks trained with gradient descent while retaining its uncertainty estimation. Furthermore, the uncertainty estimates were proportional to the model accuracy. But, the approach has a polynomial increasing kernel matrix making it infeasible for some problems. Moreover, the approach only considered deep neural networks with fully connected layers and Relu activation functions.   

Garnelo et al. \cite{GarneloSRVRET18} presented an approach with similar ethos and introduced Neural Processes (NPs). However, instead of considering a neural network with asymptotically increasing depth and width, a deep neural network was used in place of a Gaussian distribution parametrized by a kernel function to define $p(\mathbf{f} \mid \mathbf{X})$. 

$$
\begin{aligned}
p(\mathbf{y} \mid \mathbf{f}) &= \mathcal{N}\left(\mathbf{f}, \beta^{-1} \mathbf{I}\right) \\
p(\mathbf{y} \mid \mathbf{X}) &= \int p(\mathbf{y} \mid \mathbf{f})\underbrace{p(\mathbf{f} \mid \mathbf{X})}_\text{Deep neural network} d \mathbf{f}
\end{aligned}
$$

\begin{figure}[htp]
    \centering
    \includegraphics[width=0.6\linewidth]{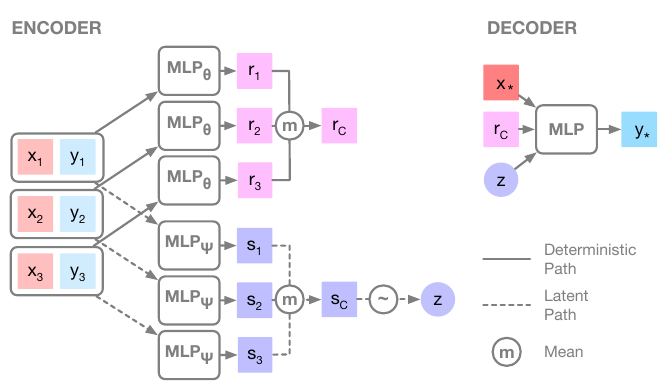}
    \caption{Neural Processes model architecture \cite{KimMSGERVT18}.}
    \label{fig:NPs}
\end{figure}

The deep neural network was trained using amortized variational inference. The consequence of such an approach is that the function space defined by deep neural networks allows us to extract hierarchical features and retain a probabilistic interpretation. The model, however, needs to be trained with Meta-learning, an approach wherein multiple diverse datasets or tasks are used to train the same model. Meta-learning is used because each function in the function space corresponds to a sequence of inputs or a task. Considering multiple tasks allows the DNN to approximate the variability of the function space. While training, a context vector $\mathbf{r}_c$ is passed to the DNN to indicate the task currently being considered as shown in Fig:~\ref{fig:NPs}. 

Moreover, to retain the probabilistic interpretation, a latent variable $z$ was introduced, which captured uncertainty in the context data. The implication being that, unlike vanilla GPs whose uncertainty comes from the kernel function and its function space, NPs do this with data. As such, the context being provided could significantly influence the model's performance and can be considered to be similar to inducing points in SGPs. 

Additionally, the model does not assume a Gaussian prior or posterior, allowing one to fit any data distribution. Garnelo et al. showed that their approach produces good predictive distributions while being parameter efficient and fast compared to a vanilla Gaussian process.

Nonetheless, the method assumed a predefined DNN model architecture which needs to be task specific. Also, the model is only an approximation to some stochastic process using a DNN. But, it isn't possible to guarantee the approximation quality of the DNN. Furthermore, the meta-learning requirement imposes a large training computation requirement, and the datasets considered have to be similar to the primary dataset of interest. 

Finally, Yang et al.\cite{YangDDS20} recently proposed an Energy-based Process (EBP). EBPs are a generalization of Neural Processes as they utilize Energy-based models \cite{lecunCHRH06} to approximate $p(\mathbf{f} \mid \mathbf{X})$ instead of a MAP trained DNN as shown below, where $f_w$ is the Energy model and $Z$ is the partition function. 
$$
p(\mathbf{f} \mid \mathbf{X}) = \frac{\exp (f_w(\mathbf{X}))}{Z(f_w)}
$$
However, by utilizing an Energy based model, the authors were able to show that vanilla GPs and NPs can be recovered from EBPs as special cases. The Energy based formulation also allows one to approximate the conditional $p(\mathbf{f} \mid \mathbf{X})$ with arbitrary distributions, unlike GP and NPs, which are limited to Gaussian distributions and the distribution defined by the DNN, respectively. 

Unlike feedforward networks which are trained to predict a label $y$ given an input  $\mathbf{X}$, Energy based models predict the Energy between a pair of $(\mathbf{X}, y)$. A well trained Energy-based will output low Energy for well matched $(\mathbf{X}, y)$ pairs and high Energy for mismatched pairs. As such, the prediction task in these models becomes a minimization task, wherein one needs to find the label $y$, which has low Energy for the given data $\mathbf{X}$. 

The consequence of training such a model to approximate our conditional in the stochastic process is that the function space is not constrained to any predefined subspace. However, Energy based models are challenging to train and require several tricks to stabilize the training process. Furthermore, it takes longer to train such models similar to models trained with Meta-learning.

\section{Discussion}
Gaussian processes have come a long way from their origins. Although many of the limitations have been addressed, there remain open problems and research directions that have not been thoroughly explored. 

One such problem is the assumption of factored output dimensions. The assumption has been made in all the methods mentioned in this paper.  It dictates that each output dimension is independent of the other. The assumption allowed factorizations that simplify some derivations, and in some cases, the assumption is required for the method to be tractable. However, the assumption might not hold in some datasets. Addressing this factorization assumption would be an interesting line of research. 

Another issue is that most SGP and DGP methods require careful model initializations and hyperparameter tuning without which, the model does not converge. Nevertheless, there aren't any formalized rules for determining model initializations and hyperparameters for most methods that could guarantee good model convergence. The tuning issue is particularly a problem when using MCMC approaches and remains to be solved. 

Additionally, MCMC methods have been shown to be successful at training DGP. But, the method need not be limited to DGP. Indeed, the technique might result in good results even for SGP. The primary motivation of using MCMC methods was to address the non-Gaussian posterior. Although vanilla GPs might not have a non-Gaussian posterior, the assumptions made in inducing methods often change this. As such, it might be worth exploring the feasibility of MCMC for training SGPs.  

Similarly, there are several variants of SG-MCMC methods that have not been benchmarked against DGPs. Havasi et al. \cite{HavasiLF18} only considered the original SGMCMC \cite{ChenFG14}  approach. However, numerous variants of the method have been introduced that improve upon SGMCMC, some of which might result in stable training dynamics.

Deep kernels in themselves were found to be highly susceptible to overfitting. However, Wilson et al. \cite{WilsonHSX16} only considered vanilla DNNs. But, DNNs could be used as Bayesian approximators as shown by Gal and Ghahramani \cite{GalG16}. Such an approach might mitigate some of the overfitting issues. 'Furthermore, one might also consider methods such as Bayes by Backprob \cite{BlundellCKW15} to train deep kernels. It would be interesting to find out the ramifications of such an approach to deep kernels. 

Garnelo et al. \cite{GarneloSRVRET18} considered a similar approach but, they consider a DNN approximation to a distribution over the function space itself. It required an explicit definition of a DNN, which needs to be task specific. Also, the model performance is dependent on the context vector to estimate prediction uncertainty which isn't consistent with a proper stochastic process. Kim et al. \cite{KimMSGERVT18} introduced a variant of Neural Processes which uses attention to improve the context vectors. Perhaps we could modify the attention mechanism to take the test data into account and generate an uncertainty estimate that incorporates test data.

Moreover, most methods assume relatively constrained function space, either with the formulation of a kernel function or a DNN. However, that need not be the case; perhaps we can consider multiple function spaces by utilizing models such as chunked hypernetworks \cite{OswaldHSG20} to generate both the model parameters and the model architecture. Thus greatly expanding the stochastic process's modeling capacity. 

Energy-based models appear to be another viable approach to expanding the function space but, the method is challenging to train and incurs substantial computational costs. Moreover, even model inference is an expensive operation requiring Hamiltonian Markov Chain methods for sampling. 

Finally, there is the issue of scalability. Although some DGP methods have been shown to scale well to large datasets, they have not been thoroughly benchmarked on highly structured datasets such as Imagenet \cite{DengDSLLF09}. The problem lies in the model depth required to achieve good performance on such a dataset. Unlike MNIST, Imagenet requires DNNs that are substantially deeper. DGPs, however, are usually only tested on models with up to ten layers. It would be indispensable to study and understand how DGPs scale to such a dataset. 
 
\section{Conclusion}
Gaussian Processes in it themselves are fascinating. Their non-parametric form, analytical properties, and ability to model uncertainty are coveted in machine learning. However, they are plagued with limitations, particularly their significant computation and storage cost. Also, conventional kernel functions restrict the family of functions that a GP could model. 

Sparse Gaussian Processes attempt to address the storage and computation cost. One dominant approach to SGPs is to use the Nyström approximation. The approach entails using variational methods to model the distribution of pseudo points for a fully Bayesian treatment. Several methods have been proposed along this line of research, each with its advantages and limitations. 

Furthermore, GPLVM was a step towards DGP. However, hierarchical feature representation was not the intended use case. It was proposed as an approach for probabilistic PCA and unsupervised learning. The Bayesian GPLVM on improved upon the original method by introducing a purely Bayesian training approach. BGPLVMs facilitated the propagation of latent space uncertainty to the posterior, thereby establishing a technique to propagate uncertainties through non-linearities in GPs.

Most of the DGP approaches considered SGPs and GPLVMs to address the issue of hierarchical feature representation. The primary trend in DGPs is to stack GPs in a feedforward manner and train them with approaches used to train SGPs and GPLVMs. However, such an approach has its limitations. The optimization bounds that were developed were not always tight, and some methods were limited to analytical solutions, which imposed scalability limitations on such techniques. 

Moreover, stacking GPs makes the model parametric as it requires a predefined model depth and layer width. Lee et al. \cite{LeeBNSPD18} considered these issues and attempted to solve them by modeling the latent function space as the space of deep neural networks. But, the approach is not feasible for real world applications as of yet and requires more work to be done to get there.

Garnelo et al. \cite{GarneloSRVRET18} consider a stochastic process parametrized with a DNN instead of a Gaussian distribution with a kernel function to define the latent function space. Still, the approach requires modeling a task specific neural network and is only an approximation to an unknown stochastic process.  Energy-based Processes address this limitation, but the method isn't mature enough as of yet. 

In conclusion, GPs are an excellent approach to model datasets. The overall trend of the field seems to be transitioning away from the Gaussian assumption and consider general Stochastic processes. The method has come a long way from its infancy, but there are still open problems that need to be addressed for it to be elevated to the prominence that it deserves. 

\bibliographystyle{apalike}
\bibliography{references}
\end{document}